\documentclass[conference]{IEEEtran}
\IEEEoverridecommandlockouts
\usepackage{cite}
\usepackage{amsmath,amssymb,amsfonts}
\usepackage{algorithmic}
\usepackage{graphicx}
\usepackage{textcomp}
\usepackage{algorithm}
\usepackage{xcolor}
\usepackage[left=1.63cm,right=1.63cm,top=1.9cm]{geometry}
\def\BibTeX{{\rm B\kern-.05em{\sc i\kern-.025em b}\kern-.08em
    T\kern-.1667em\lower.7ex\hbox{E}\kern-.125emX}}

\begin{document}

%
\title{
\huge{Efficient Cluster Selection for Personalized Federated Learning: A Multi-Armed Bandit Approach}
}

\author{
   \IEEEauthorblockN{Zhou Ni, Morteza Hashemi}

     \IEEEauthorblockA{Department of Electrical Engineering and Computer Science, University of Kansas }
     
\IEEEauthorblockA{Emails: \{zhou.ni, mhashemi\}@ku.edu}
}

\maketitle

\begin{abstract}
Federated learning (FL) offers a decentralized training approach for machine learning models, prioritizing data privacy. However, the inherent heterogeneity in FL networks, arising from variations in data distribution, size, and device capabilities, poses challenges in user federation. Recognizing this, Personalized Federated Learning (PFL) emphasizes tailoring learning processes to individual data profiles. In this paper, we address the complexity of clustering users in PFL, especially in dynamic networks, by introducing a dynamic Upper Confidence Bound (dUCB) algorithm inspired by the multi-armed bandit (MAB) approach. The dUCB algorithm ensures that new users can effectively find the best cluster for their data distribution by balancing exploration and exploitation. The performance of our algorithm is evaluated in various cases, showing its effectiveness in handling dynamic federated learning scenarios.

\end{abstract}

\begin{IEEEkeywords}
 Federated Learning, Cluster Selection, Dynamic Upper Confidence Bound.
\end{IEEEkeywords}

\section{Introduction}


Federated Learning (FL) is a decentralized learning approach to training models across multiple devices or nodes without sharing raw data. While FL offers benefits in privacy preservation and decentralized model training, it faces challenges when dealing with non-IID (non-Independently and Identically Distributed) data distributions across clients. When the number of nodes increases, FL can potentially enhance the performance of the global model. However, the scenario may differ for an individual node. If the data from one node greatly deviates from the rest in the federated network, the global model's convergence could adversely affect that particular node's learning process. As a result, participating in a non-adaptive FL network might not be advantageous. Hence, Personalised Federated Learning (PFL) emerges as a promising alternative framework. PFL, which has gained traction in recent literature \cite{huang2021personalized,li2020lotteryfl,deng2020adaptive}, seeks to adapt the training process to each client's unique data distribution. Compared to traditional FL, PFL customizes the training to the unique data profile of each device rather than having all devices contribute to a single generalized model. The models consequently grow more relevant and precise to meet the unique requirements and situations of each device.

Acknowledging the strengths of PFL, grouping users who have the same data structures as a cluster becomes a possible PFL strategy that improves the client model's overall performance. This cluster-based approach to FL aggregates local models trained on users with similar data distributions into a more representative global model. 
Hence, user clustering in FL has become a promising approach to address the challenges associated with non-IID data and personalized learning. By grouping clients with similar data distributions or other shared characteristics, user clustering can facilitate better local model training and improved model convergence, as the learning process is guided by data with higher similarity within each cluster \cite{sui2022find}. Additionally, user clustering can enhance personalization, as models trained on clustered data are more likely to capture individual clients' unique patterns and preferences \cite{briggs2020federated}. Furthermore, grouping together can reduce the communication overhead at the server side in FL, as it allows for efficient model aggregation and updating within the clusters rather than across all clients \cite{al2021reducing}. Therefore, user clustering has the potential to significantly improve the performance and efficiency of personalized federated learning systems, particularly in scenarios with non-IID data.

However, one inherent challenge is the \emph{effective grouping} of clients based on their data similarities. Current methodologies \cite{ghosh2020efficient, tan2022towards} for cluster selection often assume stationary scenarios, and they fall short in adapting to \emph{dynamic changes} in data distributions and users' participation preferences \cite{yang2023personalized} and \cite{al2023group}. For example, the user may change their data when federates in the network, or suddenly drop out from the network. Therefore, it is desirable to develop efficient clustering algorithms that achieve personalized FL under dynamic conditions.

To fill this gap, in this paper, we propose the Dynamic Upper Confidence Bound (dUCB) algorithm for cluster selection in PFL. The proposed algorithm is built upon the multi-armed bandit (MAB) frameworks, enabling dynamic decision-making under dynamic scenarios.  MAB solutions are considered as an instance of reinforcement learning (RL) solutions and are centered around the principle of balancing exploration (trying new options) with exploitation (capitalizing on known best options) to maximize cumulative rewards by selecting the optimal action among multiple available options. With its roots deeply embedded in reinforcement learning, MAB has been successfully applied across a variety of domains and has demonstrated remarkable utility, as evidenced by studies \cite{9839030},\cite{meidani2022mab}. 

In the context of FL, MAB holds great promise in addressing a myriad of challenges. The integration of MAB with FL has been recently investigated across various studies. These include, but are not limited to, resource allocation strategies \cite{azizi2022mix}, user selection methodologies \cite{xia2020multi}, \cite{larsson2021automated}, and personalization techniques \cite{wu2023fedab},\cite{shi2021federated}. In contrast to the prior works, we aim to leverage the inherent strength of MAB to achieve \emph{dynamic PFL}, so that the FL clients can dynamically select the federation cluster. This, in turn, enables the clustering algorithm to be sensitive to the dynamic conditions imposed by the users and/or the environment. 
This allows us to not only adaptively and efficiently allocate resources but also improve the performance of individual node learning models. It also has the potential to further enhance the effectiveness and adaptability of distributed learning systems, particularly in situations where data distribution and network conditions vary among clients.

In summary, our main contributions can be summarized as follows:
\begin{itemize}
    \item We consider the problem of personalized FL using clustering, and introduce a dynamic clustering solution utilizing Multi-Armed Bandit (MAB) approaches. The proposed algorithm is called dynamic Upper Confidence Bound (dUCB), which is built upon the classical UCB algorithm \cite{gonccalves2015upper, 9413628}. 

    \item Our dUCB algorithm employs a \emph{reward function} that balances exploration and exploitation, such that new FL users can identify the optimal cluster for their specific local learning model and data structures.


    \item We numerically investigate the performance of our proposed algorithm, and our results confirm that the FL clients can effectively select the optimal cluster to join. This improves the learning efficiency and system adaptability under dynamic conditions. 
\end{itemize}

The rest of this paper is organized as follows. In Section \ref{sec:problem}, we present the system model and problem formulation. In Section \ref{sec:ducb}, we introduce the proposed algorithm for dynamic cluster selection by FL clients. Section \ref{sec:results} presents our numerical results, and Section \ref{sec:conclusion} concludes the paper.

\section{System Model and Problem Formulation}
\label{sec:problem}
In this section, we present the system model, formulate the associated problems, and introduce our algorithm to address these challenges mentioned above.

\subsection{System Model}
We consider a set of users, denoted as $M = \{1, 2, ..., N\}$. Each user $i \in [M]$ consists of local data $D_i = (\textbf{x}_i, \textbf{y}_i)$ with $n_i$ samples. The users draw their input $\textbf{x} \in \mathcal{X}_{i}$ and output $\textbf{y} \in \mathcal{Y}_{i}$.  In addition, the user draws the true model parameter $\theta_i\sim\mathcal{D}_{\theta}$ from a distribution $\mathcal{D}_\theta$. The users participate in the FL cluster intending to collaboratively train a better model while retaining their data locally. 

\begin{figure}[t]
\centering
\includegraphics[width=0.5\textwidth]{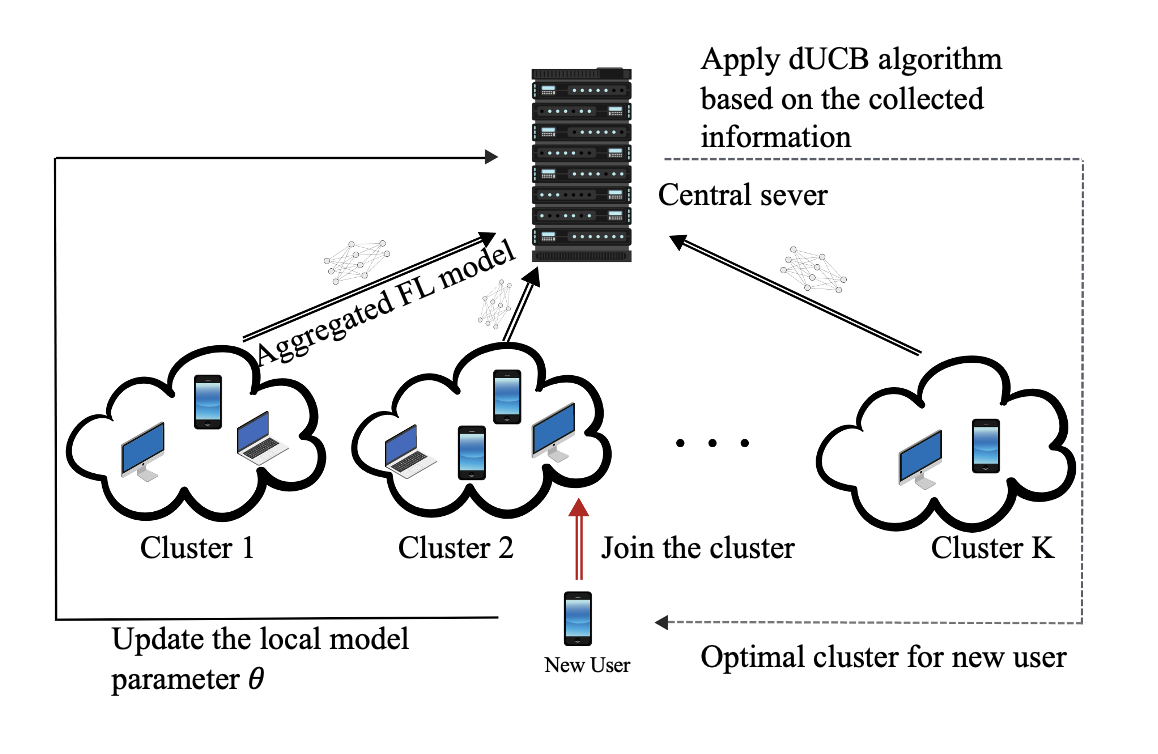}
\caption{An illustration of the system model. When a new user joins the network, it will send its local learning model to a central server. After all clusters' (cluster 1 to K in the figure) model and the new user model are collected on the server side, the cluster selection dUCB will be implemented. finally, the optimal cluster index will be returned to the new user for federated learning.}
\label{Fig:model}
\end{figure}

One of the main challenges in the FL clustering problem is that if two clients $i$ and $j$ have vastly different data distributions, forcing them to collaborate within the same cluster will likely result in worse performance than local training without collaboration.
Therefore, We employ the dynamic Upper Confidence Bound (dUCB) algorithm to address this challenge for clustering clients in an FL setting. Our method aims to find the optimal cluster for a new user, such that the performance of the FL model within each cluster improves the most when a new user joins the cluster. Figure \ref{Fig:model} depicts the system model that we consider in this work, which consists of $K$ clusters upon the arrival of a new user. The goal of the user is to achieve personalized FL by selecting the optimal cluster. 

In our system architecture, as new users enter the network, they promptly transmit their local model to the central server within the initial communication rounds. At the server end, models from every cluster and new users are aggregated, followed by the execution of the cluster selection algorithm. After determining the optimal cluster, this information is communicated back to the new users, allowing them to integrate into the most fitting cluster for FL.

\subsection{Problem Formulation}
We formulate the clustering problem in the FL context as a multi-armed bandit problem. The goal is to find the best cluster for a new user to join, which will maximize the improvement in the performance of the new user's model while federated with the optimal cluster.

\subsubsection{Federated Learning Model}

Considering personalized learning,  when a new user joins the system, the objective is to find the optimal cluster for the new user to minimize its loss in the FL process. In optimal circumstances, when the new user joins the optimal cluster, both the user's model and the global model of the cluster will be improved.
The number of users in different clusters may vary. Each user maintains a local model parameter $\theta_i$ and shares it with other users in the federation cluster.

The federated model for $\hat{\theta}_{i}^{f}$ of user $i$ can be represented as a weighted combination of the local model parameters of all users in one federation cluster:
\begin{equation}
\hat{\theta}_{i}^{f} = \sum_{j = 1}^{N_k}v_{ji}\theta_{j}, 
\end{equation}
where $v_{ji}$ is the weight assigned to user $j$'s local model by user $i$. The weights $v_{ji}$ can be chosen based on various factors, such as the number of samples contributed by each user or the importance of each user's local data. In addition, $\sum_{j = 1}^{N_k}v_{ji} = 1$ and $N_k$ denotes the number of users in the $k$-th cluster.

Our goal is to find the optimal cluster for the new users and minimize its loss:
\begin{equation}
\min \frac{1}{D_{i}}\sum_{i = 1}^{D_i}f(\hat{\theta}_{i}^{f}; \textbf{x}_i, \textbf{y}_{i}), 
\end{equation}
where $f(\theta_{i}; \textbf{x}_i, \textbf{y}_{i})$ is the loss function of user $i$, which measures the discrepancy between the federated estimation and the user's true parameter. Therefore, the new user is to find the best cluster to minimize its FL loss and improve its model performance as well.

\section{Dynamic Upper Confidence Bound Algorithm}
\label{sec:ducb}
Next, we introduce the cluster selection algorithm. Let $C = \{1, 2, ..., K\}$ denote the set of $K$ clusters available for the new user to join. We associate each cluster $k \in C$ with an arm in the multi-armed bandit framework. In each round, the server contains the new user's model and simulates it chooses a cluster to join, which is analogous to pulling an arm. The reward obtained from this action is the improvement in the performance of the global model within the chosen cluster.

We define the performance of the model as a function of its parameters, $\Theta \in \mathbb{R}^p$, and the user data size, $S_i$. The performance function can be expressed as $g(\Theta, S_i)$, where a higher value of $g$ indicates better performance. The objective is to maximize the expected improvement in the performance of its model and also the cluster model after a user joins the optimal cluster:
\begin{equation}
\max \ \mathbb{E}\left[\Delta g(\Theta, S_i)\right] = \mathbb{E}\left[R_{ik}(t)\right],
\end{equation}
where $R_{ik}(t)$ is the reward of the user $i$ for joining the cluster $k$ at time $t$, which can be defined as:
\begin{equation}
R_{ik}(t) = g(\Theta, S_{k} \cup S_{i}) - g(\Theta, S_{k}),
\end{equation}
where $S_{k}$ represents the aggregated data size of all users in the cluster $k$, and $S_{i}$ represents the local data size of user $i$. A higher value of $R_{ik}(t)$ indicates that adding user $i$ to cluster $k$ results in a greater improvement in the performance of the global model in cluster $k$. The dUCB algorithm aims to find the optimal cluster for each user that maximizes the expected reward, which is given by:
\begin{equation}
    k_i^*(t) = \underset{k \in C}{\operatorname{argmax}} \left[R_{ik}(t) + \alpha \sqrt{\frac{\log(t)}{N_{k}(t)}}\right],
\end{equation}
where $k_i^*(t) $ is the chosen cluster at time $t$, $N_{k}(t)$ is the number of times the cluster $k$ has been selected, and $\alpha$ is a tuning parameter that controls the level of exploration.

The dUCB algorithm is a variation of the classical UCB algorithm to achieve personalized FL. 
The key idea behind dUCB is to maintain an upper confidence bound on the expected improvement in performance for each cluster and to iteratively select the cluster with the highest UCB value. Algorithm \ref{alg:ducb} presents the implementation of the dUCB algorithm. 

\begin{algorithm}
\caption{Dynamic Upper Confidence Bound (dUCB) Algorithm}
\label{alg:ducb}
\begin{algorithmic}[1]
\REQUIRE $N$ users, $K$ clusters, time horizon $T$, exploration parameter $\alpha$. Initial $N_{k}(t) \gets 0$ and $R_{ik}(t) \gets 0$.
\FOR{$t = 1, \ldots, T$}
\STATE New user $i$ calculates its local model $\hat{\theta}_i$ at time $t$.
\STATE Existing user in each FL cluster computes the FL model $\hat{\theta}_{i}^{f}$.
\FOR{$k = 1, \ldots, K$}
\STATE Calculate dUCB value $I_{ik}(t) = R_{ik}(t) + \alpha\sqrt{\frac{2\log{t}}{N_{k}(t)}}$
\ENDFOR
\STATE Find the optimal cluster for user $i$ at time $t$: $k_i^*(t)  = \underset{k \in C}{\operatorname{argmax}} \ [I_{ik}(t)]$.
\STATE Update the visit counter for the selected cluster: $N_{k}(t) \gets N_{k}(t) + 1$
\STATE Update the reward function obtained from selecting cluster $k_i^*(t)$.
\ENDFOR
\RETURN Optimal cluster assignment for each user.
\end{algorithmic}
\end{algorithm}

\section{Experimental Results}
\label{sec:results}
In this section, we evaluate the performance of the proposed dUCB algorithm in a federated learning setting using synthetic datasets.

\subsubsection{Dataset}
We used a synthetic dataset, generated in a manner analogous to that described in \cite{marfoq2021federated}, to simulate the heterogeneous data distribution among clients. The dataset consists of $N$ users, each with a distinct set of data points $D_i$. The dataset was generated by sampling from different underlying distributions to create a realistic environment for federated learning with non-IID data. Furthermore, we applied the FedAvg algorithm to train the learning models and update the FL models to a central server. The reward function was then computed based on the difference between the new user's true model parameter and the clusters' FL models. This is measured at the central server with a test dataset.

\subsubsection{Evaluation Metrics}
We evaluated the performance of our dUCB algorithm using the following metrics:
\begin{itemize}
       \item \textbf{Regret}: The regret is the difference between the performance of the optimal cluster and the cluster chosen by the dUCB algorithm, which measures the accuracy of the dUCB's cluster selection.
       \item \textbf{KL divergence}: The KL divergence between the new user's true distribution and the selected cluster's distribution. A lower KL divergence indicates that the new user is more similar to that cluster.
       \item \textbf{New user's FL loss}: The loss function quantifies the difference between the predictions made by the new user's model and the actual data. This difference serves as an indication of how well the new user's model is trained or adapted to the FL environment.
\end{itemize}

\subsubsection{Experimental Scenarios}
We considered different experimental scenarios to evaluate the robustness and adaptability of our dUCB algorithm:

\begin{itemize}
    \item Varying the number of clients: We tested the performance of the dUCB algorithm for different numbers of clients and cluster numbers to analyze their scalability.

    \item Varying the data heterogeneity: We evaluated the performance under varying levels of data heterogeneity among clients to understand the robustness of the methods to different data distributions.

    \item Dynamic client participation: We simulated a dynamic environment where clients join the system during the federated learning process to evaluate the adaptability of the dUCB algorithm.
\end{itemize}


\subsection{Performance Analysis}
We conducted extensive simulations to evaluate the performance of our proposed dUCB algorithm in selecting the optimal cluster for a new user in a federated learning system. We compared the results in two scenarios, with 20 clusters and 50 clusters, to demonstrate the scalability and effectiveness of our approach.

\subsubsection{20 Clusters Scenario}
\begin{figure}[h!]
\centering
\includegraphics[width=0.5\textwidth]{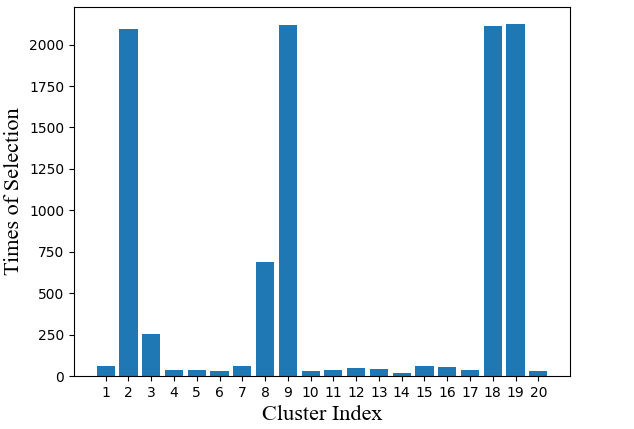}
\caption{Evaluating the number of times that each cluster was selected for 20 clusters. The 19th cluster is mostly selected which indicates it is the optimal cluster for the new user.}
\label{Fig:20-cluster}
\end{figure}

\begin{figure}[h!]
\centering
\includegraphics[width=0.49\textwidth]{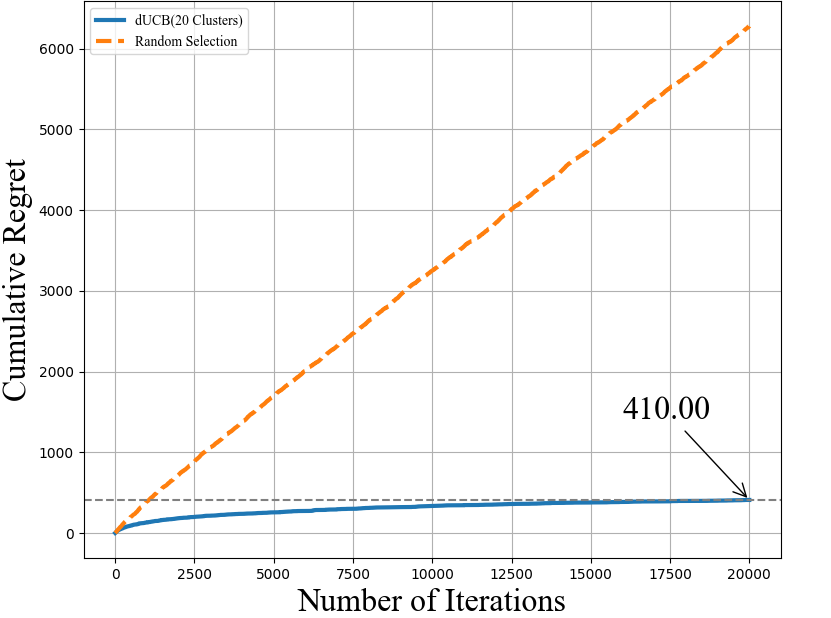}
\caption{Regret for new user with 20 clusters. We use random selection as the baseline of dUCB algorithm. After 20000 iterations, the cumulative regret converged to 410.00 which has an extraordinary performance selection accuracy compared with random selection}
\label{Figre1}
\end{figure}

We simulated a federated learning system with 20 clusters in the first scenario. After running the simulation, the algorithm converged to select those clusters with similar data distributions, as shown in Figure \ref{Fig:20-cluster}. The cumulative regret plot in Figure \ref{Figre1} exhibited an initial increase in regret, reflecting the exploration phase of the dUCB algorithm. However, as the algorithm continued to learn and perform more iterations, the regret eventually converged, demonstrating its ability to identify more suitable clusters and make better decisions. 

The performance of the dUCB algorithm can be attributed to the fact that the Kullback-Leibler (KL) divergence values were calculated for all possible cluster assignments for the new user. The KL divergence values provided a measure of the dissimilarity between the new user's model and the models of the existing clusters. In the 20-cluster scenario, as shown in Table \ref{tab:top5_min_kl_values_20}, the dUCB algorithm successfully identified the cluster with the minimal KL divergence value as the optimal cluster for the new user.

\begin{table}
\centering
\caption{Top 5 clusters with KL values of 20 clusters}
\label{tab:top5_min_kl_values_20}
\renewcommand{\arraystretch}{2} 
\begin{tabular}{| c@{\hspace{1em}} | c@{\hspace{1em}} |} 
\hline
\textbf{Cluster Index} & \textbf{KL Value} \\ 
\hline
19  & 0.0212 \\
\hline
9   & 0.0370 \\
\hline
18  & 0.1260 \\
\hline
2   & 0.2054 \\
\hline
8   & 1.8914 \\ 
\hline
\end{tabular}
\vspace{0.5em} 
\end{table}

\subsubsection{50 Clusters Scenario}
\begin{figure}[h]
\centering
\includegraphics[width=0.5\textwidth]{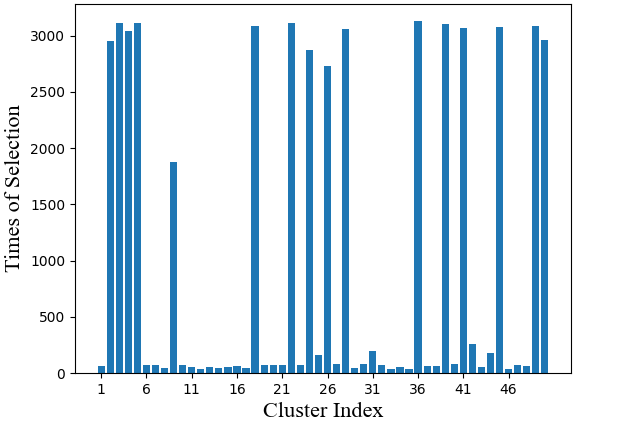}
\caption{The network size is increased to 50 clusters. The dUCB algorithm still can find the optimal cluster with the lowest KL divergence.}
\label{sim2}
\end{figure}

\begin{figure}[h]
\centering
\includegraphics[width=0.49\textwidth]{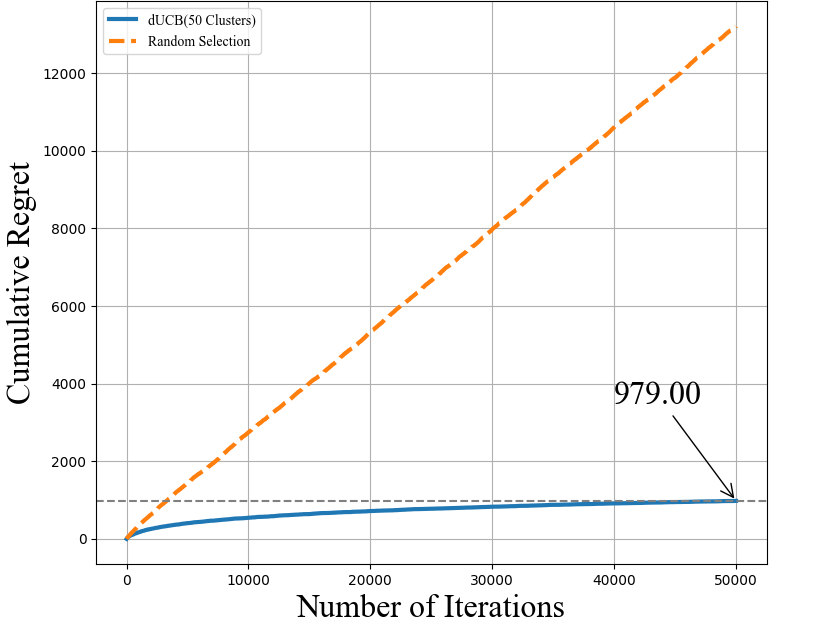}
\caption{The number of clusters is increased to 50. For such a huge system, the dUCB algorithm still can achieve a better performance with lower cumulative regret compared with random selection.}
\label{sim2reg}
\end{figure}

To further validate the performance of our proposed algorithm, we increased the number of clusters in the simulation to 50. In Figure \ref{sim2}, the dUCB algorithm was able to identify those clusters with similar distributions. This result demonstrates the scalability of the algorithm in handling larger federated learning systems with more clusters.
As shown in Figure \ref{sim2reg}, the cumulative regret plot for the 50 clusters scenario also exhibited convergence, suggesting that the algorithm was able to learn and improve its decision-making over time, even in a more complex system with a higher number of clusters.
Again, from Table \ref{tab:top5_min_kl_values}, the dUCB algorithm selected the optimal clusters for the new user with the lowest KL divergence in a large number of cluster scenarios.

\begin{table}
\centering
\caption{Top 5 clusters with KL values of 50 clusters}
\label{tab:top5_min_kl_values}
\renewcommand{\arraystretch}{2} 
\begin{tabular}{| c@{\hspace{1em}} | c@{\hspace{1em}} |} 
\hline
\textbf{Cluster Index} & \textbf{KL Value} \\ 
\hline
36  & 0.1633 \\
\hline
5  & 0.3854 \\
\hline
3  & 0.3899 \\
\hline
22  & 0.4419 \\
\hline
39  & 0.5247 \\ 
\hline
\end{tabular}
\vspace{0.5em} 
\end{table}

\subsubsection{Learning Performance evaluation}
\begin{figure}[h]
\centering
\includegraphics[width=0.5\textwidth]{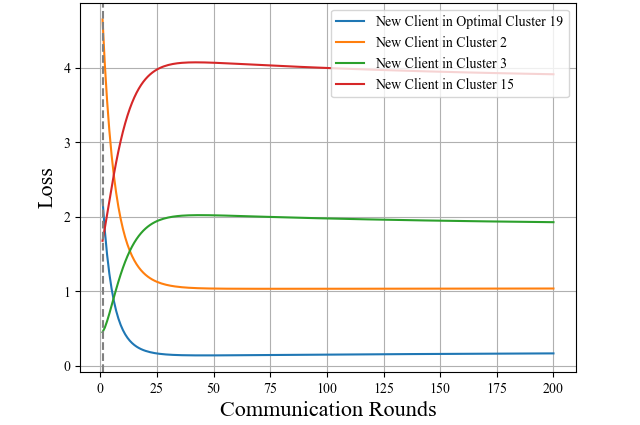}
\caption{It illustrates the difference in FL loss for a new user when it joins an optimally selected cluster versus joining other clusters. The best model performance, characterized by the lowest loss, is achieved when a new user joins the optimally selected cluster. The findings also reveal that if we force
a new user to join a cluster that deviates significantly from the new user’s
local model, the converged FL model can reduce the new user’s learning
performance}
\label{flloss}
\end{figure}

We assessed the learning performance of the new user using the dUCB algorithm, using the scenario with 20 clusters as a representative example. As shown in Figure \ref{flloss}, the new user, when joining the optimal cluster 19 as recommended by the dUCB selection algorithm, achieves minimal loss, indicating superior model performance. It's also evident that when a new user is arbitrarily assigned to a cluster (3 or 15) for FL, the resulting converged model can degrade the user's performance. Thus, in comparison to randomly selecting a cluster, our cluster selection algorithm offers users a method to join the most suitable cluster and realize enhanced performance.

\subsubsection{Summary}
The simulation results for both the 20 and 50-cluster scenarios confirm the effectiveness and scalability of the dUCB algorithm in identifying the optimal clusters for a new user in a federated learning system. The cumulative regret plots show that the algorithm converges over time, demonstrating its ability to learn and make better decisions as the simulation progresses compared with random selection. Additionally, successfully identifying the optimal clusters in both scenarios highlights the algorithm's robustness in handling different system complexities. The loss shows new user's learning performance is improved significantly by joining the optimal cluster. Overall, these results confirm that our proposed dUCB algorithm is a promising solution for selecting the optimal cluster for new users in federated learning systems, ultimately improving the overall performance and efficiency of the system.

\section{Conclusion}
\label{sec:conclusion}
In this paper, we investigated the problem of personalized FL and proposed a clustering algorithm based on the dynamic Upper Confidence Bound (dUCB) framework. The dUCB algorithm aims to find the optimal cluster for a new user
to join, maximizing the improvement in the performance of the new user. The dUCB algorithm is built upon the MAB framework such that each cluster is considered as an arm. We have shown that the dUCB algorithm can balance the trade-off between exploration and exploitation. The experimental results demonstrate the effectiveness of the dUCB algorithm in FL scenarios with relatively large numbers of clusters. There are several directions for future research to further improve the performance of the dUCB algorithm and its applicability in federated learning settings. For instance, in many federated learning applications, users may have devices with varying computation and communication capabilities. Extending the dUCB algorithm to handle heterogeneous devices can make it more suitable for a wide range of federated learning applications. By extending the dUCB algorithm to realistic FL scenarios with all the constraints across devices, data, and FL networks, our work can contribute to the development of
more efficient and scalable personalized methods for federated learning.


\section*{Acknowledgment}
The material is based upon work supported by NSF grants 1948511, 1955561, and 2212565. Any opinions, findings, and conclusions or recommendations expressed in this material are those of the author(s) and do not necessarily reflect the views of the National Science Foundation (NSF).

\vspace{16pt}

\bibliographystyle{IEEEtran}
\bibliography{ref}

\end{document}